\documentclass{article}
\PassOptionsToPackage{numbers, compress}{natbib}
\usepackage[sort]{natbib} 

\usepackage{graphicx} 
\usepackage{float} 
\usepackage{subfigure} 
\usepackage{color}
\usepackage{multirow}

\usepackage{array}
\usepackage{algorithm}  
\usepackage{algpseudocode}  
\usepackage{amsmath}  

\usepackage[utf8]{inputenc} 
\usepackage[T1]{fontenc}    
\usepackage{hyperref}       
\usepackage{url}            
\usepackage{booktabs}       
\usepackage{amsfonts}       
\usepackage{nicefrac}       
\usepackage{microtype}      
\usepackage{xcolor}         

\title{Patch-enhanced Mask Encoder Prompt Image Generation}

\author{shusong.xu, peiye.liu}

\begin{document}

\maketitle

\begin{abstract}

Artificial Intelligence Generated Content(AIGC), known for its superior visual results, represents a promising mitigation method for high-cost advertising applications. 
Numerous approaches have been developed to manipulate generated content under different conditions.
However, a crucial limitation lies in the accurate description of products in advertising applications. 
Applying previous methods directly may lead to considerable distortion and deformation of advertised products, primarily due to oversimplified content control conditions. 

Hence, in this work, we propose a patch-enhanced mask encoder approach to ensure accurate product descriptions while preserving diverse backgrounds.
Our approach consists of three components Patch Flexible Visibility, Mask Encoder Prompt Adapter and an image Foundation Model. Patch Flexible Visibility is used for generating a more reasonable background image. Mask Encoder Prompt Adapter enables region-controlled fusion. We also conduct an analysis of the structure and operational mechanisms of the Generation Module. Experimental results show our method can achieve the highest visual results and FID scores compared with other methods.

\end{abstract}

\section{Introduction}
Since its launch, Artificial Intelligence Generated Content (AIGC), a new technique in visual creation, has captured the interest of professionals in advertising and visual design. Prior to this, crafting precise product descriptions and eye-catching promotional visuals presents a significant challenge for designers, not to mention the associated high costs for advertisers.

%

As a cost-effective alternative to high-end designers, AIGC-based methods provide an automated and affordable approach to creating eye-catching promotional images through pre-trained models.
Such as the renowned example text-2-image(T2I) \cite{ramesh2022hierarchical, nichol2021glide, rombach2022high, peebles2023scalable}, have emerged as a popular choice for synthesizing images under the given human instructions. The transformer diffusion\cite{peebles2023scalable} and the latent diffusion\cite{ramesh2022hierarchical} and its extension \cite{podell2023sdxl} provided well pre-trained foundations
and improved quality and stability. Very recently, some work \cite{huang2023composer,zhao2024uni,ye2023ip,zhang2023adding} explored controllable T2I diffusion models adapting to different tasks. ControlNet\cite{zhang2023adding} and T2I-adapter\cite{zhao2024uni} 
directly incorporate adapters (or extra modules) into frozen T2I diffusion models to enable additional condition signals. 
Those works greatly enhance the quality and controllability of the image synthesis, which well demonstrate their enormous potential in advertising graphics applications.

Towards that, Some works\cite{yang2024new} used T2I model to generate background images while keeping the main product information unchanged for the advertising scene. Indirectly, some works can be used for advertising image generation.
The Uni\cite{zhao2024uni} applied 7 local conditions(e.g. Canny edge[]) and 1 global condition extracted from a reference content image. The Composer\cite{huang2023composer} also controls the output by combining CLIP\cite{ramesh2022hierarchical} sentence embeddings, image embeddings and color palettes. IP-adapter\cite{ye2023ip} provided an effective and lightweight adapter to achieve image prompt capability for the pretrained text-to-image diffusion models.
Nonetheless, current studies primarily concentrate on the generalizability of outcomes, which often falls short of meeting the demands for precise descriptions in advertising content. 
As a result, employing these methods directly in advertising graphics applications may lead to serious repercussions.
In the worst-case scenario, the mismatch between the real products and the advertising content can bring about legal ramifications. 

The main reasons for imprecise descriptions in advertising content, as observed in previous methods, can primarily be attributed to oversimplified content control conditions including prompt conditions, structure conditions and image conditions. 
For prompt conditions, the necessity of prompt engineering under the prompt-only condition demands extensive experience and numerous attempts.
Structure-based conditions, e.g., canny-edge, depth map, segmentation map, etc., usually lead to a monotonous background. This is due to the diverse backgrounds requiring a more extensive diffusion process, which may degrade the precision of content preservation.
On the other side, the image-based condition presents a more suitable solution. 
Firstly, it establishes the stylistic direction of the generated content, ensuring alignment with a predetermined aesthetic framework.
Secondly, it enhances the background detail, thereby elevating the overall visual fidelity of the advertising content.
However, there will be hallucinating results generated if the selected image contains other advertising content, as the last column images shown in figure \label{xiaoguotu}. 

Addressing these challenges, we adopt masked prompts in conjunction with visual encoders to generate background images that comply with specific requirements while preserving precise foreground details. Specifically, we introduce a flexible, patch-based module designed to segment foreground from background areas, aiming to retain adjacent style visual information as comprehensively as possible. Subsequently, we develop a masked auto-encoder for our text-image adapter, which effectively separates the visual information pertaining to the foreground and background. Moreover, the masked prompts are engineered to enhance the foreground objects by filling in missing details related to illumination, depth, boundary, and color. Finally, we incorporate a depth image of the specified object as a hint for the module, ensuring the preservation of accurate boundary information for the target object.

\begin{figure}[htbp]
    \centering
    \includegraphics[width=0.9\linewidth]{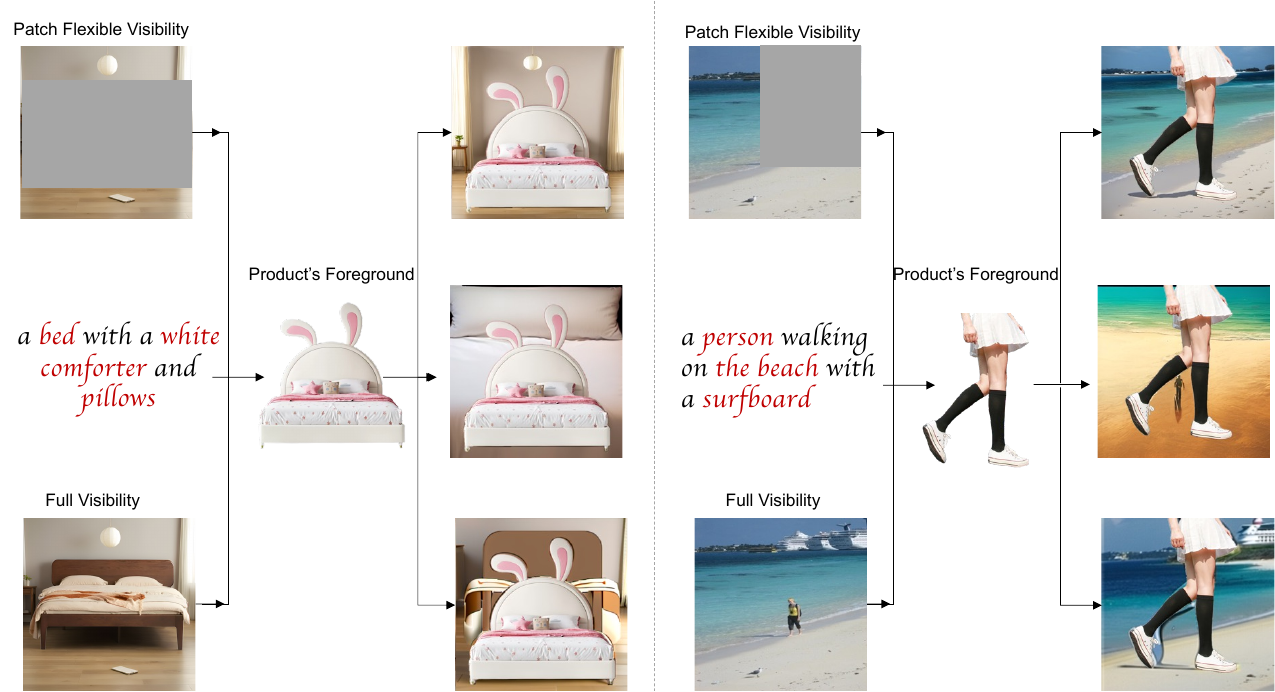}
    \caption{Various prompts for advertising image synthesis are utilized in our proposed method. These include the PFV (Patch-Enhanced via Flexible Visibility) image prompt, text prompt, and full image prompt. These prompts guide the generation of backgrounds for the advertisements. The left column demonstrates the synthesis of bed product advertising images, while the right column displays the synthesis of sock advertising graphics.}
    \label{fig:xiaoguotu}
\end{figure}

\section{Related works}
\subsection{Diffusion Models.}
\textbf{Unconditional Diffusion Models.} Diffusion models \cite{2019Generative,2021Score,2021Cascaded} are a type of generative models that produce data from Gaussian noise via an iterative denoising process. 
The currently common frameworks for diffusion
models include denoised diffusion probabilistic models
(DDPMs)\cite{sohl2015deep, ho2020denoising}, score-based stochastic differential equations (Score SDEs)\cite{song2019generative,song2020score}, conditional diffusion models\cite{rombach2022high,dhariwal2021diffusion,ho2022classifier}, etc. 
DDPMs are built around a
well-defined probabilistic process via dual Markov chains
that consist of two parts: a diffusion (or forward) process and a denoising
(or reverse) process.
Score SDE further generalizes DDPM’s discrete system to a
continuous framework based on the stochastical differential
equation\cite{song2020score}. 
Thereafter there have been numerous representative studies\cite{ma2024sit,
peebles2023scalable,2019ACN,luo2023latent,lu2023dpmsolver,zhang2023gddim,song2022denoising,liu2023genphys,xu2023pfgm}, e.g. DiT\cite{peebles2023scalable} 
aim to improve the efficiency and performance via in data process, network structure design, sampling processes, and other aspects.

\textbf{Conditional Diffusion Models.} 
In an endeavor to augment the efficacy of our model, multiple conditions have been assimilated, bifurcating the control mechanism into two distinct dimensions: firstly, control over the primary subject of the product, and secondly, modulation of the background in terms of its stylistic attributes and richness. The former dimension leverages the capabilities of the ControlNet model\cite{zhang2023adding}, facilitating precise adjustments to the input imagery to maintain the integrity of the product's primary subject. This aspect is of paramount importance within the advertising domain, wherein any deviation in the product's form or coloration is impermissible. ControlNet synergizes with image generation models such as Stable Diffusion, employing a suite of control mechanisms including but not limited to, Canny control, depth control, and mask control, thereby enabling nuanced manipulation of the image's primary subject area. Given its focus on the structural integrity of the main subject, ControlNet's functionality is categorized as local control.The latter dimension pertains to the regulation of the background's stylistic essence and richness, accomplished through the deployment of the IP-Adapter model\cite{ye2023ip}. The IP-Adapter aims to curate the background content's style and detail, facilitating the creation of images that are both aesthetically pleasing and harmonious. Utilizing a cross-attention mechanism, the IP-Adapter meticulously processes image features, generating embeddings from the supplied image and text content. These embeddings serve as dynamic parameters influencing the generative process of extensive diffusion models. The control exerted by the IP-Adapter, focusing on the stylistic delineation of the image, is thus delineated as global control.


\textbf{Image Harmonization.} 
Image harmonization is a crucial concept in the field of image composition, especially when it comes to generating images.  The essence of image harmonization resides in guaranteeing a visual and stylistic consonance throughout the entirety of the generated image. This necessitates a seamless integration of all constituents within the generated image, ensuring that no segment conspicuously deviates in terms of chromaticity, texture, or illumination. This principle attains paramount importance in the context of advertising image generation. During the image generation, it becomes imperative not only to achieve a visually cohesive narrative but also to preserve the integrity of the subject's morphology, maintain textural uniformity, and align the fabricated scene meticulously with the pre-defined input conditions.

Confronted with the intricacies of image generation assignments that demand the assimilation of multifarious conditional inputs, the attainment of image harmonization represents a formidable challenge. Recently,two methodologies stand out due to their novelty.The first approach involves the use of classifier guidance\cite{li2023image} \cite{huang2023composer}  within Latent Diffusion Models (LDM) or Denoising Diffusion Implicit Models (DDIM) for image generation. Another strategy is the concatenation of image conditions. This method involves directly appending additional image information into the model's input conditions, serving as an explicit condition for the generation process. 

\subsection{Flexible Patches}
Flexible Patches field has been received increasing attention with the popularity of models such as the Vision Transformer (ViT)\cite{dosovitskiy2020image}and the Diffusion Transformer \cite{peebles2023scalable} family. Those models falls short when dealing with images of varying resolutions though excelling within certain resolution ranges. 
Typically, input images are resized to a fixed square aspect ratio and then split into a fixed number of patches. Recent works have explored alternatives to this paradigm: FlexiViT\cite{beyer2023flexivit} supports multiple patch sizes within one architecture, enabling smooth variation of sequence length and thus compute cost. This is achieved via random sampling of a patch size at each training step and a resizing algorithm to allow the initial convolutional embedding to support multiple patch sizes. Pix2Struct \cite{lee2023pix2struct} introduced an alternative patching approach which preserves the aspect ratio, which is particularly useful for tasks such as chart and document understanding. NaViT\cite{dehghani2024patch} packs multiple patches from different images into a sequence, which enables variable resolution while maintaining aspect ratio. FiT \cite{lu2024fit} conceptualizes images as sequences of dynamically-sized tokens enables a flexible training strategy that effortlessly adapts to diverse aspect ratios during both training and inference phases, thus promoting resolution generalization and eliminating biases induced by image cropping.
\subsection{AIGC in Advertisement Production Tasks}
The AIGC technology is widely applied in the task of generating advertising images. In the previous works, creative generation methods in the advertising scene use deep learning to generate some objects/tags\cite{vempati2020enabling, mishra2020learning}, dense captions\cite{gao2022caponimage} or layout information \cite{inoue2023layoutdm,qu2023layoutllm} on image. CG4CTR\cite{mishra2020learning} use  use diffusion model to generate background images while keeping the main product
information unchanged in creative generation task for the advertising scene. In the experimental analysis, we found that modifying
the background while keeping the visual pixels of the main product
unchanged can also significantly improve the Click-Through Rate.

With the emergence of various generative AI applications, artificial intelligence-generated content (AIGC) demonstrates positive potential for design activities. However, few scholars have proposed a practical AIGC-based design methodology. This paper introduces an AIGC-empowered methodology for product color-matching design. 

An AIGC-empowered methodology to product color matching design
Language Guidance Generation Using Aesthetic Attribute Comparison for Human Photography and AIGC
Effect of AI Generated Content Advertising on Consumer Engagement
Semantic communications for artificial intelligence generated content (AIGC) toward effective content creation
Exploring User Acceptance of Al Image Generator: Unveiling Influential Factors in Embracing an Artistic AIGC Software.

\section{Approach}

\subsection{Overall}
As illustrated in figure\label{framework}, our framework consists of three main components: the Patch Flexible Visibility, the Mask Encoder Prompt Adapter, and the Foundation Models. The Patch Flexible Visibility module plays a crucial role in determining the stylistic attributes of the background image, while ensuring overall stylistic coherence throughout the composition. It enables us to seamlessly integrate different styles across the entire image. The Mask Encoder Prompt Adapter offers region-controlled fusion for multiple prompts, allowing for precise control over the integration of various prompts.
Simultaneously, the Foundation Models are trained to enhance the synergistic coherence between the foreground and its corresponding background. This ensures a harmonious integration of multi-layered components, resulting in a cohesive and visually appealing composition.

\begin{figure}[ht!]
\centering
\includegraphics[width=1.\linewidth]{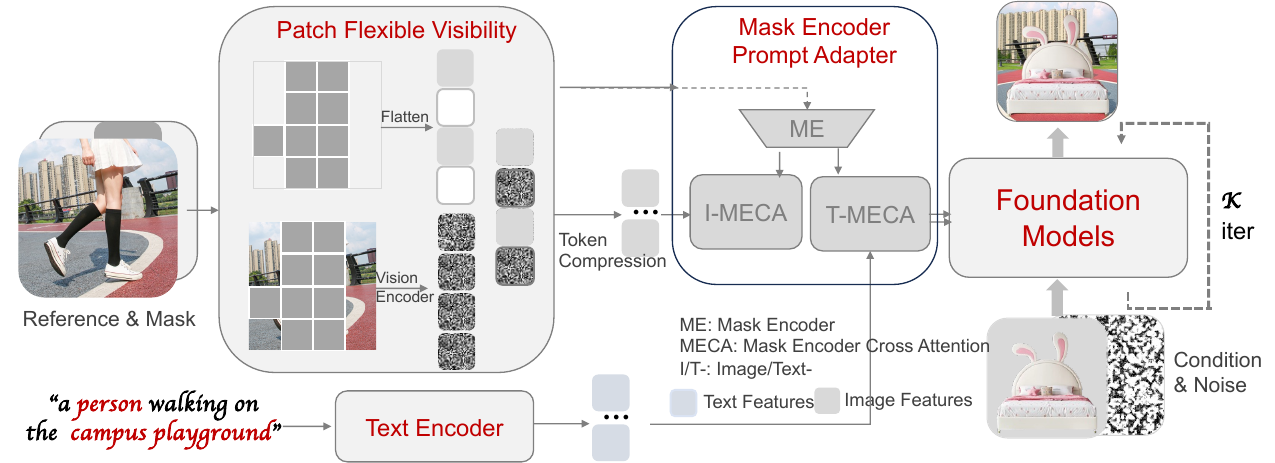}
\caption{The overall architecture of our proposed method. The patch flexible visibility masks both the reference image and its associated features. The mask encoder prompt adapter(MEPA) is consists of mask encoder(ME), text/image mask encoder cross attention(T/I-MECS). The MECA technique is employed in the fusion of text and image prompts for region control. During training, the weights of  T-MECA layers are kept frozen, ensuring their stability and consistency.}
\label{framework}
\end{figure}

\subsection{Patch Flexible Visibility}
The Patch Flexible Visibility module generates a more reasonable background image by controlling the visibility of the selection area. To achieve this, we maintain a binary decision mask $D \in \left\{ 0,1 \right\} ^ {H\times W }$ and the reference image $I_{ref} \in R ^ {H\times W \times C}$ to indicate whether to drop (the regions with $D$ value of 0) or maintain ($D$ equals 1) each element in the image encoding operation. The$H$ and $W$ are the height, width, the channel numbers.
Firstly, we re-binarized $D$ as $D_p$ based on the patch-level. we divide $D$ into $16\times 16$ patches and compute the count of 0 in each patch. If the count of 0 in each patch exceeds a certain threshold value, then all values in the current patch are set to 0. 
The we process reference image as the \cite{dosovitskiy2020image} and sent into vision encoder to compute image features.  

For image encoding, we use the encoder in CLIP\cite{dosovitskiy2020image} to better integrate it with the following text. The PFV embed masked $I_{ref} \in R^ {H \times W \times C} $ as $I_{ref}^p  \in R^ {N \times (P^2 \times C)} $, where $N=\frac{W}{P} \times \frac{H}{P}$, and then project patches as fellow:  
\begin{equation}
z_{ref} = Projection(I_{ref}^p * D_p ) * D_z
\end{equation}
where,  $z_{ref} \in R ^ {N \times Proj\_size } $, $D_z$ are flattened from re-binarized mask  $D_p$ and padded as 
$D_z \in \left\{ 0,1 \right\} ^ {N \times Proj\_size } $.
$D_p$ are flattened from the binary mask and further assignment are as follows: 
\begin{equation}
D_z[i,P^2:Proj\_size]_{ \left\{0 \leq i < N \right\} } = D_z[i,P^2]
\end{equation}
$ *$ represents element-wise matrix multiplication.

\textbf{Token Compression.}
Token Compression aims to improve information compactness and reduce hardware resource overhead. In this module, we adopt adaptive Key-Value Cache Compression\cite{ge2023model} for token compression:
\begin{equation}
Z_{c\times D}^{'} =  Q_{c \times D } \cdot Z_{ N \times D }^T \cdot Z_{N \times D}
\end{equation}
To maintain simplicity, certain linear operations have been omitted from the equation. $Q$ is a learned parameter, and $D$ represents the dimensionality of the features. The symbol $\cdot$ represents matrix multiplication. It should be noted that $c$ is a value smaller than $N$. $T$ represents matrix transpose.

\subsection{Mask Encoder Prompt Adapter}
The Mask Encoder Prompt Adapter(MEPA) enables region-controlled fusion of multi-type prompts and operates by Mask Encoder Cross Attention(MECA) in the diffusion iterative process. 

\textbf{Text/Image-Cross Attention.}
The image and text features come from the CLIP are integrated into the pre-trained diffusion model (e.g., DiT, SDXL) by the cross-attention as follows:
\begin{equation}
A = Softmax(\frac{ QK_{text}^T } {\sqrt{d}}) V_{text} + \lambda \cdot Softmax( \frac{QK_{image}^T}{\sqrt{d}}) V_{image}
\end{equation}
Where $Q=AW_q$ is the query of the attention operation. The $K_{text}=C_{text}W_{kt} $ and $K_{image}=C_{image}W_{ki}$ are the key from the text features and image features respectively. $V_{text}=C_{text}W_{vt} $ and $V_{image}=C_{image}W_{vi}$ are the value from the text features and image features. All of the W-represents are the weight matrices of the trainable linear projection layers.$\lambda$ is weight factor, and the model becomes text-to-image diffusion model if $\lambda=0$. 
Since the $W_{kt}$ and  $W_{vt}$ from text are trained well in the original prompt-based foundations, we can initial the $W_{ki}$ and  $W_{vi}$ from  the $W_{kt}$ and  $W_{vt}$ and freeze the weights of  
the text features.
here $d=h \times w \times c$ with $h$, $w$, and $c$ symbolizing the height, width, and channel count for each segment, respectively. 

\textbf{Mask Encoder Cross Attention.}
The Mask Encoder Cross Attention enables region-controlled cross attention through masked queries. The queries have the same shape as the noise latent. The noise latent is obtained by gradually adding noise to the latent space vectors obtained from encoding $I_{ref}$ using VAE\cite{kusner2017grammar}. Subsequently, the VAE latent is passed into the foundation models. Specifically, in our approach, we employ the SDXL\cite{podell2023sdxl} foundation model.
To ensure query matching, we utilize the VAE and the foundation model structure with all parameters set to 0, including weights and offsets. The query\_mask is then incorporated as one of the inputs to the adapter, alongside text embedding and image embedding, within the diffusion process.
\begin{equation}
MA = zeros\_{SDXL}(zeros\_{VAE}(D) )
\end{equation}
The Mask encoder for Adapter are re-represented:
\begin{equation}
A = Softmax(\frac{ ((1-MA) \times Q)K_{text}^T } {\sqrt{d}}) V_{text} + Softmax( \frac{(MA \times Q)K_{image}^T}{\sqrt{d}}) V_{image}
\end{equation}

\subsection{Generation Module}
In this subsection, we present an analysis of the structure and operational mechanisms of the Generation Module, which is primarily based on diffusion models. The main purpose of this module is to ensure coherence in the synthesis of foreground and background, enabling a seamless integration of various components. To achieve this, we utilize Local Control Foundation Models\cite{zhang2023adding} as the designated models for detailed management. The inputs for this module comprise a masked reference depth map and a foreground depth map. The module generates the final advertisement image that meets the predefined requirements.
Diffusion models are generative frameworks that involve two sequential processes: the diffusion process, also known as the forward process, which introduces Gaussian noise to the data in a systematic manner through a Markov chain with $T$ steps; and the denoising process, where a model with learnable parameters reconstructs samples starting from Gaussian noise. It is important to mention that these models can be conditioned on additional inputs. In the context of a diffusion model, the training objective is formally defined as $\boldsymbol{\epsilon}_{\theta}$, a functional predictor of noise, which represents a streamlined variant of the variational boundary:
\begin{equation}
\label{eq:trainingObjective}
L_{\text{simple}} = \mathbb{E}_{\boldsymbol{x_0}, \boldsymbol{\epsilon}}\sim \mathcal{N}(\mathbf{0}, \mathbf{I})|\boldsymbol{\epsilon} - \boldsymbol{\epsilon}_\theta(\boldsymbol{x}_t, \boldsymbol{c}, t)|^2,
\end{equation}
wherein $\boldsymbol{x}_{0}$ represents the authentic dataset augmented with an auxiliary condition $\boldsymbol{c}$; $t\in [0, T]$ specifies the temporal juncture within the diffusion trajectory; $\boldsymbol{x}_t = \alpha_t\boldsymbol{x}_0+\sigma_t\boldsymbol{\epsilon}$ epitomizes the noised data at the $t^{th}$ interval; and $\alpha_t$, $\sigma_t$ are pre-established functions of $t$ that orchestrate the diffusion narrative. Post-training of the model $\boldsymbol{\epsilon}_{\theta}$, this framework enables the iterative generation of images from stochastic noise inputs.

\section{Experiments}
\subsection{Experimental Settings}
\textbf{Datasets.} 
The SAM-1B\cite{kirillov2023segment} and COCO\cite{lin2014microsoft} datasets serve as our training datasets. SAM-1B comprises 11 million diverse, high-resolution images that prioritize privacy protection, along with 1.1 billion high-quality segmentation masks collected by Meta. On the other hand, COCO is a dataset that includes 91 common object categories, with 82 of these categories having more than 5,000 labeled instances. 
For evaluation, we selected 50 images from the COCO test dataset, comprising 5000 images, and another 50 images from the advertising images dataset as references. These references guided the generation of 5000 samples for product foregrounds. The generated samples were then used to compute similarity with the COCO test dataset of 5000 images, as well as with another dataset of 5000 advertising images. 
Additionally, we collected a dataset of 100,000 advertising images for both training and Evaluation purposes. 

\textbf{Comparison method.}
To demonstrate the effectiveness of our method, we compare it with other existing methods. The comparisons are categorized into three types: text-to-image methods with only text prompts: SDXL\cite{podell2023sdxl} and Pixel-a\cite{chen2023pixart}; image-only prompts: Uni-Control\cite{zhao2024uni} and IP-Adapter\cite{ye2023ip}; and text and image prompts Uni-Control\cite{zhao2024uni} and IP-Adapter\cite{ye2023ip}.

\textbf{Evaluation Metric.}
To assess the performance of our proposed method, we employ three distinct evaluation metrics:
\textbf{FID} (Frechet Inception Distance), which quantifies the similarity between the original product images and the images generated by the model. It assesses the accuracy and diversity of the generated images in relation to the main subject of the product.
\textbf{IS} (Inception Score), which is calculated by inputting generated images into a pre-trained Inception-v3 model and extracting predicted class probabilities. The IS score is derived from the average Kullback-Leibler (KL) divergence between the class distribution of generated images and a reference dataset. A higher IS score indicates higher quality and diversity in capturing various modes of the target distribution. This metric is essential for evaluating GAN performance and guiding optimization.
\textbf{MoS}(Mean of score), which refers to the average value of a set of scores or ratings. It is calculated by summing all the scores and dividing the sum by the total number of scores.
Besides, we conducted a qualitative evaluation of our method by selecting images of various types and styles. In the case of the model that solely relies on text prompts, we input the reference image into the BLIP \cite{li2022blip} subtitle model to generate corresponding text descriptions.

\textbf{Implementation Detail.} 
Our experiments were conducted using the SDXL pre-trained model\cite{podell2023sdxl}, with OpenCLIP ViT-H/14 \cite{dosovitskiy2020image} being utilized as the image encoder. The model was trained for 1 million steps on a single machine equipped with 8 V100 GPUs, with a batch size of 8 per GPU. The AdamW optimize was employed, with a fixed learning rate of 0.0001 and a weight decay of 0.01.
During the training process, we resized the images to have the shortest side equaling 512 pixels, and then performed center cropping to achieve a resolution of 512 × 512 pixels. Random masks were generated for the reference images. In the inference stage, these masks could be manually selected or obtained through the SAM framework.
To expedite model convergence, we included a $5\%$ probability of utilizing only text prompts. When image prompts were used, there was a $50\%$ probability of applying masks. During the inference stage, we utilized a DDIM sampler with 30 steps and set the guidance scale to 7.5.

\begin{table}[t]
  \caption{Comparison with Existing Methods}
  \label{sample-table}
  \centering
  \begin{tabular}{llcccccc} 
    \toprule
    Type & Method    & Text   & Image   & Region-controlled & FID(ad)\(\downarrow\)  & FID(coco)\(\downarrow\) \\
    \midrule
    Text-only & Control-dit-text &  \(\checkmark\) & \(\times\) &\(\times\) & 283.04 & 348.49   \\
    & control-sdxl-text & \(\checkmark\) &\(\times\)&\(\times\)&253.10 & 332.38 \\
    \midrule
    Img-only & Uni-ControlNet-img & \(\checkmark\) & \(\checkmark\) &\(\times\) &275.88& 330.13\\
    & IP-Adapter-plus-img& \(\checkmark\) & \(\checkmark\) &\(\times\) &277.80&347.02\\
    & Ours-img& \(\checkmark\) &\(\checkmark\) &\(\checkmark\)&237.70&317.44\\ 
    \midrule
    Text-Img & Uni-ControlNet-img+text& \(\checkmark\) &\(\checkmark\) &\(\times\) &253.42& 324.60\\
    & IP-Adapter-plus-img+text& \(\checkmark\) &\(\checkmark\) &\(\times\) &255.9& 349.73\\
    & Ours-img+text& \(\checkmark\) &\(\checkmark\) &\(\checkmark\) &\textbf{233.37}&\textbf{317.13} \\
    \bottomrule
  \end{tabular}
\end{table}


\subsection{Comparison with Existing Methods}
We compared 3 types generation models. The models "Pixel-a" \cite{chen2023pixart} and "SDXL" \cite{podell2023sdxl} operate under the control of text prompts. "Uni-I" and "Uni-I-T" represent the results of \cite{zhao2024uni} under text prompts and text-image prompts. "IP-I" and "IP-I-T" represent the results of \cite{ye2023ip} under text prompts and text-image prompts. The text used for these models is converted from the "Ref" image using BLIP \cite{li2022blip}.
Quantitative comparison are showed as table\label{sample-table} and qualitative comparison are showed as figure\label{resall}. 

\begin{figure}[ht]
\centering
\includegraphics[width=1.0\linewidth]{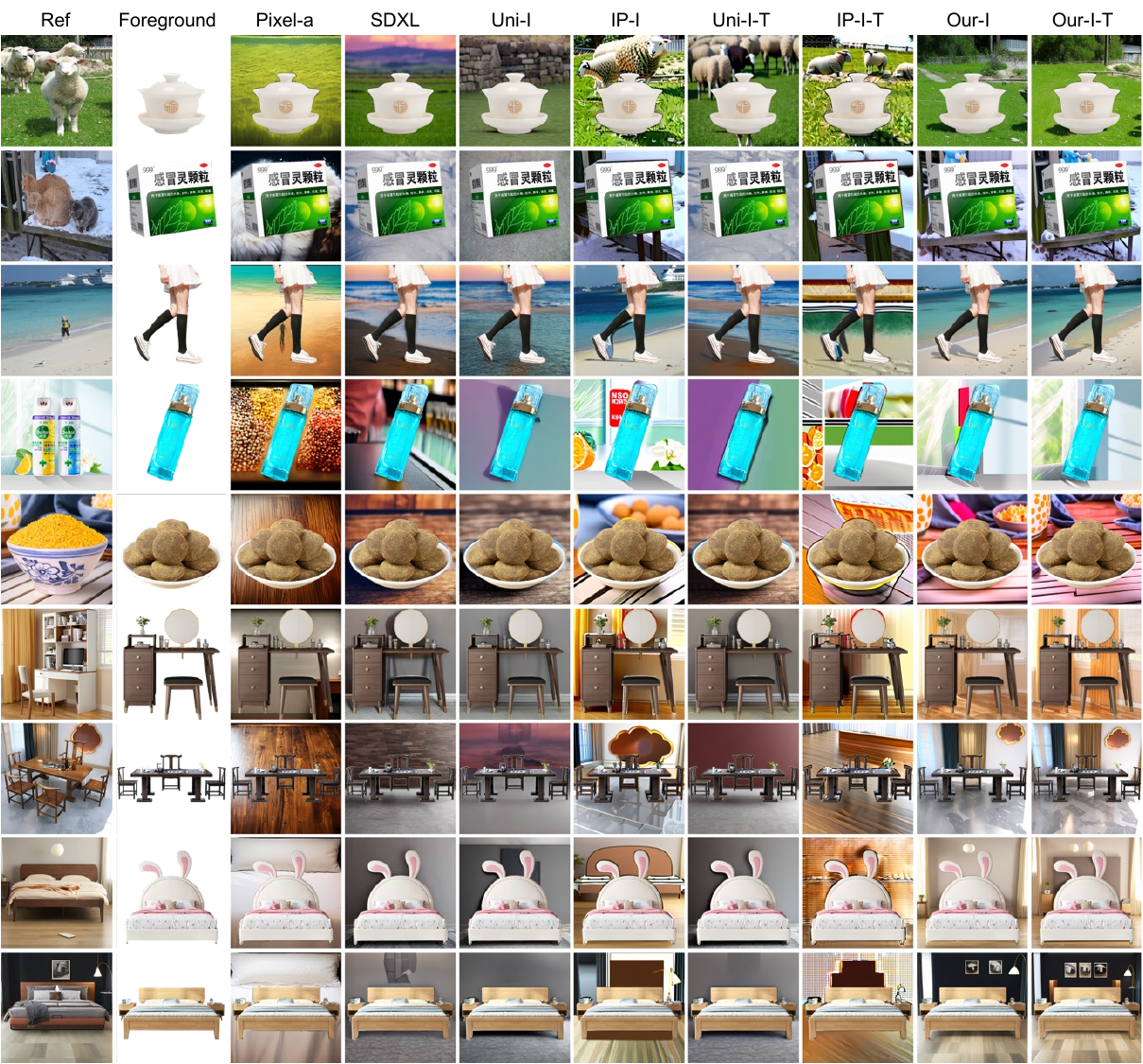}
\caption{Comparisons. "Ref" refers to the reference image that guides the style, while "The foreground" represents the product. The models "Pixel-a" \cite{chen2023pixart} and "SDXL" \cite{podell2023sdxl} operate under the control of text prompts. "Uni-I" and "Uni-I-T" represent the results of \cite{zhao2024uni} under text prompts and text-image prompts. "IP-I" and "IP-I-T" represent the results of \cite{ye2023ip} under text prompts and text-image prompts. The text used for these models is converted from the "Ref" image using BLIP \cite{li2022blip}.}
\label{resall}
\end{figure}

\subsubsection{Quantitative Comparison}
From the table \ref{sample-table}, our method named "Ours-img+text" and "Ours-img" achieved the best metrics. FID(ad) represents the distance between the synthesized advertisement images and the user-preferred advertisement images, while FID(coco) represents the distance between the generated advertisement images and real coco dataset. Our FID values are the lowest, indicating that our method, which only requires image input, achieves excellent results slightly below the image-text method.
\subsubsection{Qualitative Comparison}
In Figure \ref{resall}, it can be observed that the method solely based on text generation often lacks harmony with the foreground (columns three and four). Conversely, the method based on image generation may suffer from inappropriate background textures. In contrast, our method excels in generating backgrounds that closely resemble the reference image while seamlessly blending the foreground and background regions. It effectively avoids any irregular texture artifacts and offers natural boundaries, such as shadows and lighting effects. Overall, our method outperforms both qualitatively and quantitatively, establishing its superiority over alternative approaches.

\subsection{Ablation Study}
\subsubsection{Effects of the Patch Flexible Visibility}
To validate the effects of Patch Flexible Visibility (PFV), we conducted ablation experiments comparing the full reference image (without PFV). The figure\label{pfv} demonstrates the superiority of PFV.
In this ablation experiment section, the figure\label{pfv} illustrates the qualitative results. It is evident that our method effectively avoids foreground textures appearing in the newly generated images, resulting in a more coherent overall outcome. However, when the entire image is used as input, there can be issues with foreground blending and unnatural elements, leading to unsatisfactory generated advertisement images that do not meet clients' expectations. Additionally, we calculated the FID and MoS scores, which are detailed in the appendix. The FID score is smaller and the MoS is higher with PFV.
\begin{figure}[ht]
\centering
\includegraphics[width=1.0\linewidth]{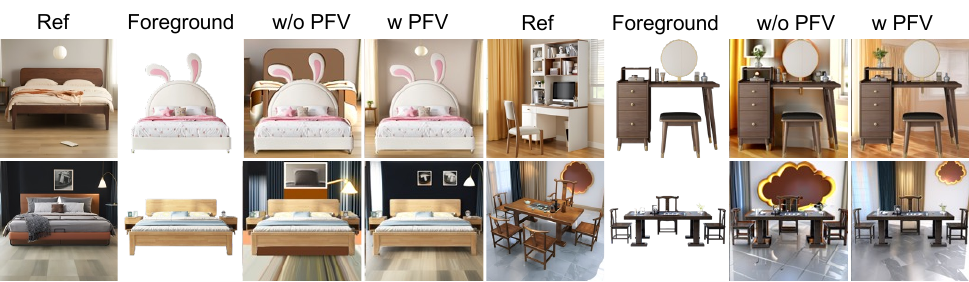}
\caption{Effects of the Patch Flexible Visibility. "Ref" refers to the reference image that guides the style, and "The foreground" represents the product. "w/o" indicates the condition without RFV (Reference Image Guided Style), while "w" represents the condition with RFV. }
\label{pfv}
\end{figure}

\subsubsection{Effects of the Mask Encoder Prompt Adapter}
To validate the effectiveness of Mask Encoder Prompt Adapter (MEPA), we compared the methods of global weighted fusion(named "Global prompts" in figure\label{maskencoder}), image-reference only(named "I-prompt"), and mask encoder fusion(named "ME prompt"). Both qualitative as showed by figure \label{maskencoder} and quantitative comparisons(detail in the appendix) were conducted. The qualitative results are depicted in the figure, demonstrating that global weighted fusion can lead to blurred backgrounds. This is because the global fusion technique combines the text-generated image with the original reference image. Although there is some alignment, ensuring identical textures between the two is challenging, resulting in a loss of background spatial composition. Directly inputting the image may exhibit slightly inferior performance in certain details compared to the local fusion method. Furthermore, subjective evaluations of the generated images were conducted. The local fusion method and reference image input garnered higher scores compared to global fusion. The local fusion method obtained slightly higher scores than the reference image input.

\begin{figure}[ht]
\centering
\includegraphics[width=1.0\linewidth]{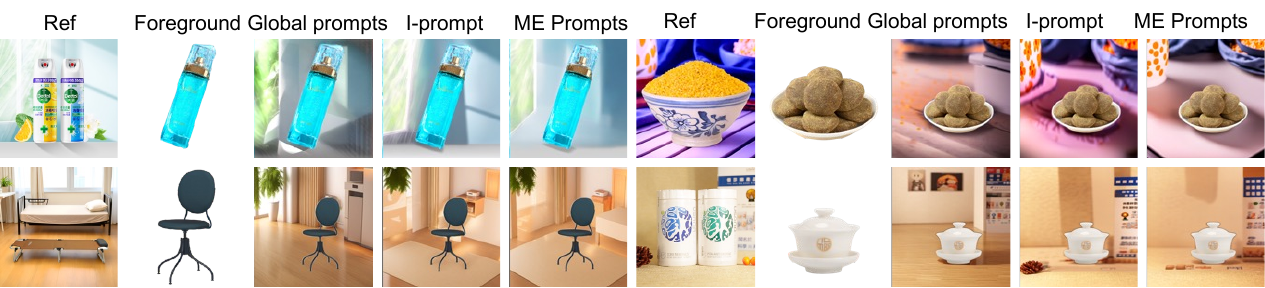}
\caption{Effects of the mask encoder prompt adapter. 
 "Ref" refers to the reference image that guides the style, while "The foreground" represents the product. "Global prompts" indicate the global weights assigned to the image prompt and text prompt. "I-prompt" denotes the results achieved solely under the image prompt. "ME prompts" represent the mask encoder prompts generated by our MECA mentioned above.}
\label{maskencoder}
\end{figure}

\section{Conclusion}

In this paper, we present an innovative method for the generation of advertising imagery. The architecture of our model is divided into two distinct modules: the Art Control Module (ACM) and the Generation Module (GM). The ACM is tasked with defining the stylistic aspects of the image's background as well as ensuring the stylistic consistency of the entire generated image. The GM undertakes the task of generating the background, facilitating a seamless integration between the product layers and the background layers. In Art Control Module, in order to eliminate undesired content from reference images, we propose the implementation of the Patch Flexible Visibility (PFV) module and Mask Encoder Prompt Adapter(MEPA). Experiments across various datasets have proven the superiority of our approach over previous methods, both in preserving the authenticity of original product details and in creating high-quality images. The empirical evidence, whether quantitative or qualitative, demonstrates that our method outperforms existing fine-tuned models and adapters.

\section*{References}

\medskip





\bibliographystyle{unsrtnat}

\bibliographystyle{plain}
\bibliography{ref}

\end{document}